\documentclass[pdflatex,sn-apa]{sn-jnl}

\usepackage{graphicx}%
\usepackage{multirow}%
\usepackage{amsmath,amssymb,amsfonts}%
\usepackage{amsthm}%
\usepackage{mathrsfs}%
\usepackage[title]{appendix}%
\usepackage{xcolor}%
\usepackage{textcomp}%
\usepackage{manyfoot}%
\usepackage{booktabs}%
\usepackage{algorithm}%
\usepackage{algorithmicx}%
\usepackage{algpseudocode}%
\usepackage{listings}%
\usepackage{tabularx,array,threeparttable}
\lstnewenvironment{prompt}[1]
  {\medskip\noindent\textit{#1}\endgraf\nopagebreak}
  {\medskip}
\theoremstyle{thmstyleone}%
\theoremstyle{thmstyletwo}%

\theoremstyle{thmstylethree}%

\begin{document}

\title[LLM Judges and Examinees for Item Difficulty]{LLMs as Judges and Examinees for Medical Item Difficulty Prediction}


\author*[1]{\fnm{Xiaomeng} \sur{Xiong}}\email{xiaomengxiong@ufl.edu}

\author[2]{\fnm{Marcus} \sur{Walker}}

\author[2]{\fnm{Yanlin} \sur{Jiang}}

\affil*[1]{\orgname{University of Florida}, \orgaddress{\city{Gainesville}, \state{FL}, \country{USA}}}

\affil[2]{\orgname{National Commission on Certification of Physician Assistants}, \orgaddress{\country{USA}}}


\abstract{\mathversion{normal}Predicting medical item difficulty from item text alone remains challenging. Prior work has largely relied on linguistic features or text representations derived from general-purpose and domain-specific language models, but predictive performance has remained modest. This study takes a different approach by using difficulty-related signals that emerge when open-weight large language models (LLMs) interact with test items. Using USMLE and PANCE medical licensure examination items, we examined LLMs in two roles. As judges, the models provided absolute difficulty ratings and pairwise difficulty comparisons. As simulated examinees, they generated response-based signals including correctness, keyed-option probability, response entropy, and the probability margin between the top two options. We then combined the strongest judgment and response signals with baseline item features in downstream prediction models. Pairwise difficulty judgment was the most consistent individual LLM signal across both datasets and outperformed absolute ratings, and adding simulated-examinee signals yielded only small further gains. The best configurations reached a Spearman correlation of $\rho = .447$ on USMLE, exceeding the best system in the 2024 BEA shared task ($\rho = .25$), and $\rho = .494$ on PANCE. These results were comparable to direct absolute judgments from the proprietary frontier model Claude Opus 5 ($\rho = .443$ and $.488$, respectively). Overall, judgment and response signals from open-weight LLMs captured difficulty-related information beyond text representations. Because these models can be deployed locally, the approach offers a practical option for predicting the difficulty of unreleased or high-security test items.}


\maketitle

\section{Introduction}
In educational measurement, item difficulty is fundamental to practices such as test assembly, equating, item bank maintenance, and computerized adaptive testing (CAT). Traditionally, the difficulty of newly developed items is estimated through pretesting using examinee response data. However, pretesting often requires substantial time and resources and depends on collecting a sufficient number of examinee responses. In high-stakes testing, it can also increase the risk of exposing unreleased items \citep{alkhuzaey2024text, maeda2025field}. Estimating the difficulty of new items before historical response data become available therefore remains an important practical challenge. Even imperfect difficulty predictions may have practical value. Within an item response theory framework, auxiliary information about items can be incorporated into parameter estimation to improve the precision and stability of item calibration when examinee response data are limited \citep{mislevy1988exploiting}. Such information can also supplement or even replace examinee response data when linking new test forms to an established scale \citep{mislevy1993equate}.

In recent years, researchers have attempted to predict the difficulty of medical examination items directly from item text, but performance has remained limited. In the BEA 2024 shared task on medical item difficulty prediction, the best system achieved a Spearman correlation of .25, a Pearson correlation of .27, and an RMSE of .299 \citep{gombert2024predicting, yaneva2024findings}. \citet{ha-etal-2019-predicting} used a dataset nearly 18 times larger, containing 12,038 USMLE items, yet their full-feature model reached only $r = .32$.

One major line of work in this area has focused on learning difficulty-related representations from item text. Approaches have included word embeddings and linguistic features \citep{ha-etal-2019-predicting}, general-purpose embedding models \citep{ram2024leveraging, yousefpoori2024using}, end-to-end fine-tuned Transformer models \citep{felice2024british, gombert2024predicting, li2025item, rodrigo2024uned}, and language models pretrained on medical corpora \citep{bulut2024item, fulari2024utilizing, tack2024itec}. Despite their technical differences, these approaches share a common modeling strategy. They learn a representation from item text and then map that representation to empirical item difficulty. This line of work implicitly assumes that a sufficiently informative text representation can capture information related to item difficulty.

However, the difficulty of a medical examination item is not simply a function of its linguistic complexity or the medical content it covers. High-stakes medical examinations typically follow strict item-writing guidelines that discourage unnecessary material, overly complex stems and options, misleading information, and linguistic cues that may reveal the correct answer \citep{ha-etal-2019-predicting}. These practices improve item quality, but they also reduce surface-level linguistic variation that might otherwise serve as a proxy for difficulty. At the same time, items covering similar medical topics or semantic content may still differ substantially in empirical difficulty. Further improvements in text representation alone may therefore have limited returns. This motivates the central idea of the present study: Rather than relying on representations extracted from item text and mapping them directly to difficulty, can more informative difficulty-related signals emerge when large language models interact with items through judgment or actual response behavior?

We examine this question from two complementary perspectives. First, in the LLM-as-a-judge role, models directly evaluate the absolute difficulty of an item or compare the relative difficulty of two items. Second, in the LLM-as-an-examinee role, models answer the items and provide difficulty-related signals through their responses and option probability distributions. The former reflects how a model evaluates item difficulty, whereas the latter reflects how the model behaves and how uncertain it is when attempting to solve the item. These two sources may capture different aspects of difficulty.

We systematically examine difficulty signals generated through LLM-item interactions using two independent medical licensure examination datasets, USMLE and PANCE. Six open-weight models are used as difficulty judges to produce both absolute ratings and pairwise judgments. Pairwise judgments are aggregated into a common relative difficulty scale using a sparse comparison graph and the Bradley-Terry model. Thirteen open-weight models are also used as simulated examinees, from which we extract four response-based measures: correctness, keyed-option probability, response entropy, and the probability margin between the top two options. Finally, judgment-based signals, response-based signals, and baseline item features are combined in downstream prediction models to evaluate their incremental and complementary value.

The results show that pairwise judgment is the most consistent individual LLM difficulty signal across the two datasets. Adding simulated-examinee signals to pairwise judgments yields small further gains in both datasets, although these gains are within sampling uncertainty. The best configuration achieves a Spearman correlation of $\rho = .447$ on USMLE, exceeding the approximately .25 reported for the best BEA 2024 shared-task system using the same official data split. On PANCE, the best configuration reaches $\rho = .494$. The combined signals from open-weight models were also comparable to direct absolute difficulty judgments from the frontier commercial model Claude Opus 5 on both datasets. This has particular practical relevance for high-stakes testing. Unreleased or highly secure examination items are often subject to strict security and data governance requirements, making it potentially infeasible to send item content to external commercial APIs. Locally deployable open-weight models therefore offer particular practical value in this setting.

The study addresses the following research questions:

RQ1. When large language models serve as item difficulty judges, how does judgment quality vary across models and judgment formats, including absolute judgment and pairwise comparison? For pairwise comparisons, how do true difficulty gaps and comparison density further affect judgment performance?

RQ2. When large language models serve as simulated examinees, which models and response-derived signals provide the most informative indicators of item difficulty?

RQ3. To what extent do judgment-based and response-based signals provide complementary information for predicting item difficulty across two medical licensure examination datasets?

\section{Related Work}
\subsection{Medical Item Difficulty Prediction}
The first public benchmark for medical item difficulty prediction came from the shared task organized by NBME at the BEA workshop in 2024 \citep{yaneva2024findings}. Before this, because exam data is difficult to share, work in this area remained fragmented across individual institutions, with no transparent evaluation on a public dataset and no meaningful basis for comparing approaches. The task released 667 retired USMLE clinical items and asked participants to predict item difficulty (proportion correct) from item text and metadata alone. The winner \citep{gombert2024predicting} ranked first among the 43 difficulty submissions made by 17 teams (RMSE = 0.299, $r = .27$, $\rho$ = .25), yet improved only slightly over a dummy regressor predicting the mean (RMSE = 0.311, ranked 16 out of 43).

\citet{gombert2024predicting} attributed part of the difficulty of this task to the small dataset and the highly specialized domain. However, \citet{ha-etal-2019-predicting} worked with nearly 18 times as many items (12,038 USMLE Clinical Knowledge MCQs) to predict difficulty (proportion correct), and their full-feature model still reached only $r = .32$. They drew on three classes of features: embeddings (Word2Vec and ELMo), 113 linguistic features, and information retrieval features that quantified how difficult an item was for an automatic question-answering system retrieving over MEDLINE. The highest correlation between any individual feature and difficulty was only about .11, and surface complexity in particular failed. The mechanistic explanation offered by \citet{ha-etal-2019-predicting} is important, and it is a matter of design rather than accident. These USMLE items were written by experienced item writers following a common set of guidelines, which required avoiding verbosity, ``window dressing'' (extraneous material not needed to answer the item), ``red herrings'' (information designed to mislead the test-taker), overly long or complicated stems and options, and various grammatical cues (correct answers that are longer, more specific, or more complete than the other options, or the inclusion of the same word or phrase in both the stem and the correct answer). The authors acknowledge that this practice, while sensible, makes modeling difficulty very challenging.

In other words, we have reason to hypothesize that, in medical items, item-writing guidelines deliberately minimize the surface-level variation that might otherwise serve as a proxy for difficulty. As a result, the items are designed to differ primarily in their textual content. We further hypothesize that the semantic content of an item is not equivalent to the difficulty of that content. Much of the work conducted both during and after the shared task has followed the same general approach: using representation learning to extract the semantics of the item text and attempting to map those semantic representations directly onto item difficulty. This includes end-to-end fine-tuned transformer encoders \citep{gombert2024predicting, rodrigo2024uned, felice2024british} and decoders \citep{li2025item}, semantic vectors extracted with general-purpose embedding models \citep{ram2024leveraging, yousefpoori2024using}, and models pre-trained on medical corpora such as Bio\_ClinicalBERT, BiomedBERT, and PubMedBERT \citep{tack2024itec, bulut2024item, fulari2024utilizing}. The assumption behind this line of work is that difficulty is encoded in the item text in some deeper semantic form, and that a sufficiently good representation will read it out. But the official summary \citep{yaneva2024findings} reports that even the best solution outperformed the baseline (predicting the mean) by only a small margin. Medical domain adaptation performed worse. In a same-architecture comparison, the winning team found that BiomedBERT and BiomedELECTRA fell short of both their general-purpose counterparts and the dummy baseline \citep{gombert2024predicting}. These results suggest that stronger domain-specific semantic representations alone may not be sufficient to recover empirical item difficulty.

Therefore, these findings suggest that further improvements in textual representation alone may be insufficient. Representation learning can capture rich semantic and linguistic information from item text, but these representations do not necessarily isolate the aspects that determine empirical difficulty. Surface-level variation is constrained by standardized item-writing practices, while items with similar medical content can still differ substantially in difficulty. Thus, richer text representations may still leave important difficulty-related variation unexplained.

The limited performance of text representations does not imply that LLMs cannot provide useful information about item difficulty. Difficulty-related information may emerge from how LLMs interact with exam items, beyond what can be captured by text-based representations alone. These signals can come from two complementary perspectives. An LLM can serve as a judge with broad medical knowledge, comparing the relative difficulty of different items. It can also serve as a simulated examinee, with its response performance indicating how difficult an item is for an artificial agent. We treat these two sources of information as complementary proxies for difficulty and examine whether they can capture difficulty-related information that is missed by text representations alone.

\subsection{LLM Pairwise Comparison}

Pairwise comparison has a long history in psychological and educational measurement, dating back to Thurstone's law of comparative judgment \citep{thurstone1927law}. In educational assessment, comparative judgment has been used to evaluate student work by asking judges to compare pairs of responses and aggregating these judgments into a common scale \citep{pollitt2012method}. More recently, LLMs have been used to make pairwise judgments across a range of tasks.

\citet{qin2024large} reframed information retrieval ranking as a pairwise decision task, asking an LLM to determine which of two candidate documents is more relevant to a query. Their Pairwise Ranking Prompting approach, based on moderately sized open-source LLMs, outperformed supervised ranking baselines across seven BEIR benchmarks and achieved average gains of more than 10\% over pointwise LLM approaches on standard ranking metrics.

A similar pattern has been observed in difficulty prediction. \citet{raina2024question} compared supervised task-transfer methods, LLM-based absolute difficulty estimation, and LLM comparative prompting on 658 reading-comprehension MCQs. Zero-shot pairwise comparisons with ChatGPT-3.5 achieved a Spearman correlation of .404, substantially higher than the .183 obtained from absolute difficulty judgments with the same model and slightly higher than the .387 achieved by a supervised level-classification model trained on 100,388 RACE++ questions. These results suggest that, when direct item-to-score mapping is limited, reformulating the task as a relative judgment offers an alternative way to improve difficulty ranking.

For aggregating local pairwise judgments into a global ranking, \citet{qin2024large} and \citet{raina2024question} mainly relied on win-based aggregation or sorting procedures. \citet{liusie2024efficient} compared win-ratio aggregation with the Bradley-Terry model. The two approaches performed almost identically when all comparisons were available, but Bradley-Terry showed small and consistent gains across multiple LLM judges when only a subset of pairs was observed. Unlike simple win counting, the Bradley-Terry model represents comparison outcomes through the relative latent scores of the two items and jointly estimates those scores from all observed pairwise comparisons \citep{bradley1952rank}. For a difficulty prediction task with hundreds of items, exhaustive pairwise comparison requires $O(N^2)$ comparisons and quickly becomes expensive. A sparse comparison design combined with Bradley-Terry aggregation therefore offers a more scalable way to recover a common latent difficulty scale from a limited set of pairwise judgments.

\subsection{LLM Simulated Examinees}
Whereas pairwise comparison uses LLMs to judge item difficulty, a complementary approach treats LLMs as simulated examinees and derives difficulty-related signals from their response behavior. One line of work prompts the same LLM to simulate examinees at different grade or proficiency levels \citep{benedetto2024using, srivatsa2025can, acquaye2026take, li2026can}. However, this type of proficiency simulation is sensitive to the model and prompt design, and the simulated ability levels do not always correspond reliably to the target student populations. \citet{srivatsa2025can} found that grade-enforcement prompting can change model performance, but some models fail to preserve the expected pattern of higher ability at higher grade levels. \citet{li2026can} reported a similar pattern across more than 20 LLMs. Even when models were explicitly instructed to simulate low-, average-, and high-proficiency examinees, their actual problem-solving performance did not consistently reflect these intended ability differences. High-capability models were particularly unable to suppress their existing knowledge and problem-solving ability.

To avoid relying on a single model to simulate multiple ability roles, another line of work uses heterogeneous response patterns produced by different LLMs. \citet{park2024large} used LLMs from different model families, scales, and training backgrounds to construct synthetic examinee groups at different ability levels and estimated item difficulty from their zero-shot responses. On a database course question dataset, the estimated difficulty had a Pearson correlation of .348 with IRT difficulty derived from real student responses. Other studies have generated difficulty-related performance signals by changing the conditions under which the model answers questions. \citet{duenas2024upn} progressively removed information from item stems and varied the sampling temperature of ChatGPT-3.5, then used model performance under these conditions as features for difficulty prediction. The best single feature had an absolute Spearman correlation of about .29 with observed difficulty, and the final system ranked second in the BEA 2024 NBME shared task.

Overall, these studies suggest that LLM response behavior can provide useful performance-based signals for item difficulty prediction. However, it remains unclear whether these performance-based signals provide incremental information beyond judgment-based difficulty estimates, or whether the two types of signals largely overlap.

\section{Method}
Prior work suggests that surface-level linguistic variation in medical examination items is constrained by standardized item-writing practices, while semantic content alone does not necessarily correspond to item difficulty. Consequently, directly mapping semantic representations of item text onto empirical difficulty may not fully leverage the medical knowledge encoded in large language models. Motivated by this limitation, we explore difficulty-related signals that emerge from interactions between LLMs and medical test items.

We consider two types of LLM-derived difficulty signals: judgment-based signals (LLM-as-a-judge), obtained through absolute ratings and pairwise comparisons, and response-based signals (LLM-as-an-examinee), obtained from models’ direct responses to the items. We then examine variation across models, elicitation methods, response readouts, and pairwise-comparison density, and evaluate whether combining the two signal types improves difficulty prediction.

Our primary analyses focus on open-weight models that can be deployed within an institution’s own computing environment. As an external performance reference, we additionally obtain a single absolute difficulty rating from Claude Opus 5 for each test item in both datasets, providing a comparison with a frontier commercial model. Claude Opus 5 was accessed through the Anthropic API.

\subsection{Datasets}
Two datasets were used: the NCCPA PANCE dataset and the publicly released BEA 2024 shared-task dataset from NBME \citep{yaneva2024findings}. All analyses were conducted independently for the two datasets.

NCCPA PANCE. The dataset contained 528 PANCE items. Item difficulty was represented by the b parameter estimated under a Rasch model, with higher values indicating greater difficulty. Each item included a question stem, response options, and the correct answer. The median stem length was 420 characters. Thirty-five items (6.6\%) contained an image, indicated by an \texttt{<img>} tag in the item text. This indicator served as the image feature in the baseline model. The data were split into training and test sets using a 70/30 stratified random split, yielding 370 training items and 158 test items.

BEA 2024 (NBME). The dataset contained 667 retired USMLE clinical items and used the official split of 466 training items and 201 test items. Difficulty was defined as a linear transformation of the proportion of examinees answering the item correctly, with higher values indicating greater difficulty. Each item included a question stem, response options, the correct answer, item type (text only / image-containing), and examination Step (Step 1/2/3). The median stem length was 696 characters. Seventy items (10.5\%) contained images.

\subsection{LLM-as-a-judge}
Judgment-based difficulty signals were generated using the six open-weight models listed in Table~\ref{tab:models}, selected to span differences in model size, model family, and domain-specific pretraining. Each model produced two types of difficulty judgments. All open-weight models were accessed through a university-hosted inference platform, and no item content was sent to external services. Decoding used a temperature of 0 where supported and default settings otherwise.

\begin{table}[htbp]
\centering
\caption{Open-Weight LLMs Used as Difficulty Judges and Simulated Examinees}
\label{tab:models}
\footnotesize\setlength{\tabcolsep}{2pt}
\small
\begin{tabular}{lllll}
\toprule
Model & Parameters & Architecture & Developer & Role \\
\midrule
nemotron-3-super-120b-a12b & 120.6B / 12.7B active & Hybrid MoE & NVIDIA     & Judge, examinee \\
nemotron-3-nano-30b-a3b    & 31.6B / 3.6B active   & Hybrid MoE & NVIDIA     & Judge, examinee \\
gpt-oss-120b               & 117B / 5.1B active    & MoE        & OpenAI     & Judge, examinee \\
gemma-4-31b-it             & 30.7B                 & Dense      & Google     & Judge, examinee \\
medgemma-27b-it            & 27B                   & Dense      & Google     & Judge, examinee \\
gemma-3-27b-it             & 27B                   & Dense      & Google     & Judge, examinee \\
\addlinespace
gpt-oss-20b                & 21B / 3.6B active     & MoE        & OpenAI     & Examinee \\
llama-3.3-70b-instruct     & 70B                   & Dense      & Meta       & Examinee \\
llama-3.1-70b-instruct     & 70B                   & Dense      & Meta       & Examinee \\
llama-3.1-8b-instruct      & 8B                    & Dense      & Meta       & Examinee \\
mistral-small-3.1          & 24B                   & Dense      & Mistral AI & Examinee \\
mistral-7b-instruct        & 7B                    & Dense      & Mistral AI & Examinee \\
granite-3.3-8b-instruct    & 8B                    & Dense      & IBM        & Examinee \\
\bottomrule
\end{tabular}
\end{table}

For absolute judgments, models were asked to estimate each item's difficulty percentile on a 0--100 scale, with higher values indicating greater difficulty. Each item was evaluated once in a zero-shot setting, without anchor items of known empirical difficulty. 

For pairwise judgments, models were asked to identify which of two items would be more difficult for the target examinee population. All models used the same comparison graph. Each item participated in at least 40 pairwise comparisons (40--43; mean = 40.5). Opponents were assigned by a degree-balancing procedure that repeatedly paired an item with the least-compared eligible candidate, with ties broken at random and without using empirical difficulty information. No item pair was repeated. Each pair was evaluated twice with the presentation order reversed. Consistent judgments were recorded as wins or losses, and inconsistent judgments were assigned equal weights of 0.5 to both items. This procedure yielded 13,520 unique item pairs for USMLE and 10,680 for PANCE.

Pairwise outcomes were then aggregated using a Bradley--Terry model with L2 regularization ($\lambda = 0.001$) to obtain a continuous relative difficulty score for each item. Within each dataset, training and test items were jointly calibrated on the same comparison graph so that all items were placed on a common latent difficulty scale. Because no empirical difficulty labels enter the Bradley--Terry fit, joint calibration does not use outcome information from test items. This approach is consistent with an operational scenario in which newly developed items are compared against items in an existing item bank. To examine the effect of comparison density, we additionally constructed lower-density subgraphs from the full comparison graph and refit the Bradley--Terry model at different values of $k$, allowing us to evaluate how predictive performance changed as the number of comparisons per item decreased. Full prompts for all tasks are provided in Appendix~\ref{app:prompts}.

\subsection{LLM-as-an-examinee}
Response-based difficulty signals were obtained by having the models directly answer the multiple-choice items and extracting difficulty-related features from their response distributions. We included 13 candidate examinee models, six of which overlapped with the judge models. The remaining models broadened coverage across model sizes, families, and capability levels, as shown in Table~\ref{tab:models}.

Option probabilities were derived from the log probabilities of the answer-option letters and normalized across the available options. Non-reasoning models generated a single token, with probabilities extracted from that position. Reasoning models were allowed up to 512 tokens and instructed to end with ``Answer: X'', and probabilities were extracted from the position following ``Answer:''.

Four response readouts were derived from the same model call. Binary correctness indicated whether the highest-probability option matched the keyed answer. Keyed-option probability was the normalized probability assigned to the correct option. Entropy summarized uncertainty across the option probability distribution, with larger values indicating a more diffuse distribution. Margin was defined as the probability difference between the highest- and second-highest-probability options, providing a measure of response confidence.

\subsection{Feature Selection and Model Development}
We first selected the strongest-performing source within each type of LLM-derived signal using the training set:

\begin{itemize}
    \item \textbf{LLM pairwise judgment (P):} Among the six pairwise LLM judges, we retained the model with the strongest individual performance on the training set.

    \item \textbf{LLM absolute judgment (A):} Among the six absolute-rating LLM judges, we retained the model with the strongest individual performance on the training set.

    \item \textbf{LLM examinee readouts (S):} For each of the four response readouts (keyed-option probability, binary correctness, entropy, and margin), we selected the model with the strongest training-set performance, so the four readouts in S could come from different examinee models.
\end{itemize}

The selected LLM-derived signals were combined with baseline item features ($B$; stem length, option length, number of options, image indicator, and, for USMLE only, exam step) and entered into three downstream prediction models: ridge regression, random forest, and XGBoost. Within each feature configuration, the prediction model and its hyperparameters were selected jointly on the training set by five-fold cross-validation repeated five times, using the mean cross-validated Spearman correlation as the criterion (candidate settings are listed in Table~\ref{tab:learners}). Predictors were standardized using training-set means and standard deviations. The selected model was then refit on the full training set and evaluated once on the held-out test set. We compared eight prespecified configurations (see Table~\ref{tab:feature_configurations}) to examine the incremental contribution and complementarity of the different LLM-derived signal sources, and report all eight rather than selecting a single winning configuration. Uncertainty in test-set Spearman correlations was quantified with 95\% percentile bootstrap confidence intervals based on 2,000 resamples of test items. Differences between key configurations, and between the best configuration and Claude Opus 5, were evaluated by paired bootstrap resampling of test items (2,000 resamples), with both correlations computed on each resample.

\begin{table}[htbp]
\centering
\caption{Feature Configurations Evaluated in the Downstream Prediction Models}
\label{tab:feature_configurations}
\small
\begin{tabularx}{\textwidth}{
    >{\raggedright\arraybackslash}p{2.5cm}
    >{\raggedright\arraybackslash}X}
\toprule
\textbf{Configuration} & \textbf{Included Features} \\
\midrule
$B$
& Baseline item features \\

$B + P$
& Baseline features + selected LLM pairwise judgment \\

$B + A$
& Baseline features + selected LLM absolute judgment \\

$B + S$
& Baseline features + four selected LLM examinee readouts \\

$B + P + A$
& Baseline features + LLM pairwise judgment + LLM absolute judgment \\

$B + P + S$
& Baseline features + LLM pairwise judgment + LLM examinee readouts \\

$B + A + S$
& Baseline features + LLM absolute judgment + LLM examinee readouts \\

$B + P + A + S$
& Baseline features + LLM pairwise judgment + LLM absolute judgment + LLM examinee readouts \\
\bottomrule
\end{tabularx}

\end{table}

\section{Results}
\subsection{LLM-as-a-judge }
Pairwise difficulty judgment accuracy did not increase monotonically with nominal model size (Table~\ref{tab:judges}). Gemma 4 31B achieved the highest pairwise accuracy on both USMLE and PANCE, outperforming the two models with nominal parameter counts of 120B. This pattern also highlights that total parameter count is not a directly comparable measure of model size across architectures.

\begin{table}[htbp]
\centering
\caption{Pairwise Judgment Accuracy of the Six Judge Models}
\label{tab:judges}
\footnotesize\setlength{\tabcolsep}{3pt}
\small
\begin{tabular}{lrrrcrrr}
\toprule
& \multicolumn{3}{c}{USMLE} & & \multicolumn{3}{c}{PANCE} \\
\cmidrule{2-4}\cmidrule{6-8}
Model & Acc$_{\text{cons}}$ & Acc$_{\text{all}}$ & Consistency & &
        Acc$_{\text{cons}}$ & Acc$_{\text{all}}$ & Consistency \\
\midrule
gemma-4-31b-it             & .668 & .639 & 82.6 & & .668 & .645 & 86.1 \\
nemotron-3-super-120b-a12b & .656 & .617 & 74.7 & & .653 & .614 & 74.5 \\
gpt-oss-120b               & .624 & .599 & 79.8 & & .633 & .608 & 81.2 \\
nemotron-3-nano-30b-a3b    & .603 & .573 & 71.3 & & .618 & .581 & 69.1 \\
medgemma-27b-it            & .599 & .577 & 77.5 & & .582 & .562 & 76.0 \\
gemma-3-27b-it             & .574 & .560 & 80.9 & & .537 & .531 & 82.2 \\
\bottomrule
\end{tabular}
\vspace{2mm}
\footnotetext{
\footnotesize
\textit{Note.} $\mathrm{Acc}_{\text{cons}}$ is accuracy on pairs with consistent judgments across both presentation orders. $\mathrm{Acc}_{\text{all}}$ includes all pairs, with inconsistent judgments scored as 0.5. Consistency is the percentage of pairs with agreement across presentation orders.
}
\end{table}

Gemma 4 also differs from Gemma 3 in more than parameter count. It represents a newer generation of the Gemma family with a greater emphasis on reasoning, including configurable thinking capabilities and changes to its attention architecture. The strong performance of Gemma 4 in this study may therefore reflect broader improvements in model architecture and reasoning capabilities rather than simply an increase from 27B to approximately 31B parameters.

Domain adaptation may also matter. At the same 27B parameter scale, MedGemma consistently outperformed Gemma 3 across both datasets, suggesting that adaptation to the medical domain may further support judgments of medical item difficulty.

Order consistency likewise did not show a simple relationship with either model size or judgment accuracy. Gemma 4 combined the highest pairwise accuracy with the highest consistency on PANCE and relatively high consistency on USMLE. At the same time, Gemma 3 remained highly consistent despite substantially lower accuracy. These results suggest that a model may make the same judgment reliably across presentation orders without that judgment necessarily being correct.

\begin{figure}[t]
\centering
\includegraphics[width=\linewidth]{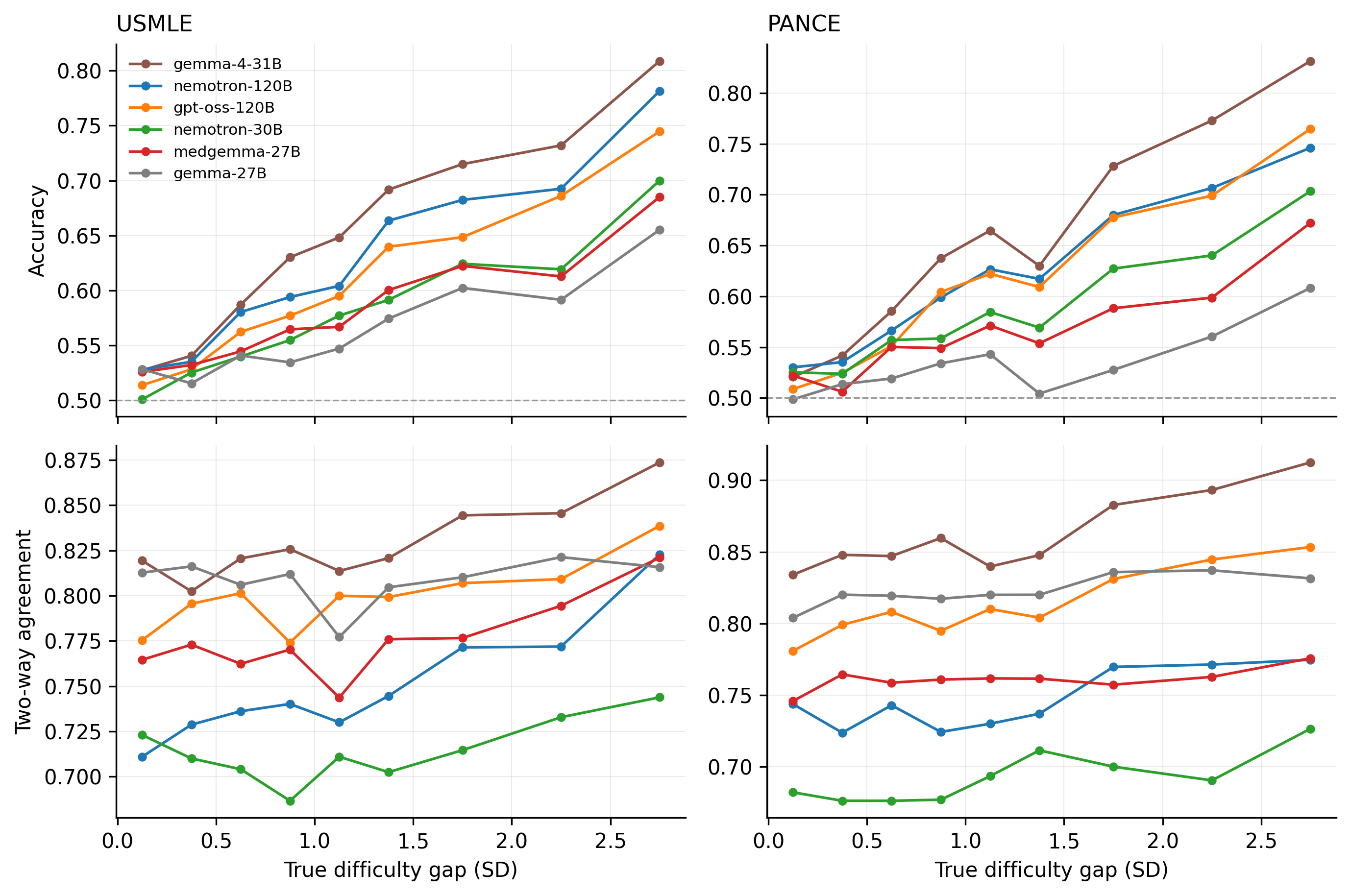}
\caption{Pairwise Judgment Accuracy (Top) and Two-Way Agreement (Bottom) as a Function of the Empirical Difficulty Gap}
\label{gap.png}
\end{figure}
As shown in Figure~\ref{gap.png}, pairwise judgment accuracy generally increased as the true difficulty gap between the two items became larger. Gemma 4 31B achieved the highest or near-highest accuracy across almost all difficulty-gap intervals and showed a clear increase with larger gaps on both datasets, exceeding .80 in the largest gap interval. The two 120B models also showed relatively clear upward trends. In contrast, the remaining smaller models generally benefited from larger difficulty gaps but showed more local fluctuations across the intermediate intervals. Gemma 3 27B showed the weakest improvement, particularly on PANCE, where its accuracy remained close to chance at moderate difficulty gaps.
Greater separation in true item difficulty primarily made it easier for models to identify which item was harder, but did not necessarily make their judgments more stable to changes in presentation order. Two-way agreement did not show a similar systematic increase and appeared to reflect model-specific stability more than sensitivity to the magnitude of the difficulty difference.

As the number of pairwise comparisons per item ($k$) increased, the correlation between Bradley--Terry difficulty scores and empirical item difficulty generally improved (Table~\ref{tab:ksweep}), but the gains showed clear diminishing returns. Using the correlation at $k = 40$ as the full-density reference, most models reached approximately 95\% of this performance with only 10 to 15 comparisons per item.

Increasing the number of comparisons also did not compensate for weaker judgment quality. This pattern was most apparent for Gemma 3 27B. On USMLE, its correlation increased only from .158 at $k=5$ to .174 at $k = 40$, while on PANCE it remained nearly unchanged at approximately .06 across comparison densities. 
\begin{table}[htbp]
\centering
\caption{Spearman Correlations Between Bradley--Terry Scores and Empirical Difficulty Across Comparison Densities}
\label{tab:ksweep}
\small
\begin{tabular}{lrrrrrrrr}
\toprule
 & \multicolumn{8}{c}{Comparisons per item ($k$)} \\
\cmidrule(lr){2-9}
Judge & 5 & 10 & 15 & 20 & 25 & 30 & 35 & 40 \\
\midrule
\textit{USMLE} \\
gemma-4-31B & .384 & .412 & .419 & .420 & .422 & .423 & .422 & .424 \\
nemotron-120B & .349 & .377 & .390 & .389 & .392 & .394 & .395 & .396 \\
gpt-oss-120B & .297 & .328 & .330 & .338 & .337 & .341 & .341 & .342 \\
nemotron-30B & .245 & .282 & .280 & .288 & .291 & .293 & .292 & .294 \\
medgemma-27B & .210 & .222 & .226 & .230 & .229 & .232 & .231 & .232 \\
gemma-27B & .158 & .172 & .173 & .171 & .173 & .173 & .173 & .174 \\
\addlinespace
\textit{PANCE} \\
gemma-4-31B & .402 & .428 & .433 & .437 & .438 & .437 & .438 & .439 \\
nemotron-120B & .296 & .332 & .339 & .346 & .347 & .346 & .347 & .348 \\
gpt-oss-120B & .269 & .303 & .315 & .315 & .319 & .319 & .319 & .318 \\
nemotron-30B & .213 & .230 & .245 & .244 & .244 & .248 & .245 & .246 \\
medgemma-27B & .124 & .143 & .153 & .150 & .156 & .160 & .157 & .158 \\
gemma-27B & .056 & .056 & .066 & .061 & .063 & .063 & .062 & .062 \\
\bottomrule
\end{tabular}
\vspace{2mm}
\footnotetext{
\footnotesize
\textit{Note.} For each $k$, comparison graphs were randomly subsampled to at most $k$ comparisons per item, and Bradley--Terry scores were recalculated. Values are mean correlations over 20 repetitions on the training set.
}
\end{table}

As shown in Table~\ref{tab:abs_vs_pair}, the relative ordering of the models was broadly consistent across absolute and pairwise judgment. This suggests that the models’ underlying ability to judge item difficulty was relatively stable across elicitation formats. However, pairwise comparison produced stronger difficulty signals overall. The best training-set Spearman correlations from pairwise judgment were .424 on USMLE and .439 on PANCE, compared with .349 and .411 for absolute judgment.

\begin{table}[htbp]
\centering
\caption{Training-Set Spearman Correlations for Absolute and Pairwise Judgments}
\label{tab:abs_vs_pair}
\small
\begin{tabular}{lrrcrr}
\toprule
& \multicolumn{2}{c}{USMLE} & & \multicolumn{2}{c}{PANCE} \\
\cmidrule{2-3}\cmidrule{5-6}
Model & Absolute & Pairwise & & Absolute & Pairwise \\
\midrule
gemma-4-31b-it             & \textbf{.349} & \textbf{.424} & & \textbf{.411} & \textbf{.439} \\
nemotron-3-super-120b-a12b & .263 & .396 & & .249 & .348 \\
gpt-oss-120b               & .288 & .342 & & .226 & .318 \\
nemotron-3-nano-30b-a3b    & .239 & .294 & & .176 & .246 \\
medgemma-27b-it            & .134 & .232 & & .131 & .158 \\
gemma-3-27b-it             & .077 & .174 & & .154 & .062 \\
\bottomrule
\end{tabular}
\vspace{2mm}
\footnotetext{
\footnotesize
\textit{Note.} Absolute judgments are 0--100 difficulty ratings; pairwise judgments are Bradley--Terry scores obtained at $k = 40$. Boldface indicates the highest correlation in each column.
}
\end{table}

\subsection{LLM-as-an-examinee}

The pattern was different for LLM-as-an-examinee. The most informative response-based signals did not always come from the largest models (see Table~\ref{tab:examinee_readouts} in the Appendix). On USMLE, Mistral Small 3.1 24B provided the strongest keyed-option probability, entropy, and margin signals, with correlations ranging from .287 to .327, while Nemotron 3 Nano 30B had the highest correlation for binary correctness ($\rho = .270$). On PANCE, Llama 3.1 70B provided all three continuous readouts, each with a correlation above .380, whereas binary correctness was strongest for Granite 3.3 8B ($\rho = .269$).

Across both datasets, the three probability-based readouts showed higher correlations with empirical item difficulty than binary correctness. Keyed-option probability had the highest correlation on PANCE ($\rho = .386$) and tied with margin on USMLE ($\rho = .327$). Differences among the three continuous readouts were relatively small, particularly on PANCE, where their correlations ranged from .384 to .386.

\subsection{Combining LLM-Derived Signals for Difficulty Prediction}
Table~\ref{tab:signals} summarizes the best-performing signal from each source described above. For pairwise judgment, absolute judgment, and each examinee readout, the model with the largest absolute training-set Spearman correlation with empirical item difficulty was selected for subsequent combination analyses.

Across the eight feature configurations (Table~\ref{tab:feature_performance}), LLM pairwise judgment ($P$) was the most consistent individual LLM-derived signal across both datasets. Adding $P$ to the baseline features increased the Spearman correlation from .228 to .425 on USMLE and from .100 to .450 on PANCE. The surface-feature baseline on USMLE already approached the best performance reported in the 2024 BEA shared task ($\rho = .25$). LLM examinee readouts ($S$) also improved prediction. The contribution of absolute judgments ($A$) varied across datasets.
\begin{table}[htbp]
\centering
\caption{Test-Set Performance Across the Eight Feature Configurations}
\label{tab:feature_performance}
\footnotesize\setlength{\tabcolsep}{3pt}
\small
\begin{tabular}{lllrcrr}
\toprule
Dataset & Configuration & Model & Spearman $\rho$ & 95\% CI & Pearson $r$ & RMSE \\
\midrule
USMLE & B       & Ridge & .228 & [.086, .361] & .274 & .299 \\
      & B+P     & Ridge & .425 & [.301, .533] & .417 & .283 \\
      & B+A     & Ridge & .357 & [.225, .479] & .381 & .288 \\
      & B+S     & Ridge & .379 & [.250, .496] & .382 & .287 \\
      & B+P+A   & Ridge & .419 & [.292, .531] & .423 & .281 \\
      & B+P+S   & Ridge & \textbf{.447} & [.329, .551] & .441 & .280 \\
      & B+A+S   & Ridge & .411 & [.289, .523] & .427 & .281 \\
      & B+P+A+S & Ridge & .436 & [.313, .547] & .443 & .278 \\
      & Claude Opus 5 (reference) & --- & .443 & --- & .397 & --- \\
\midrule
PANCE & B       & RF    & .100 & [$-$.062, .245] & .093 & 1.159 \\
      & B+P     & Ridge & .450 & [.316, .568] & .484 & 1.018 \\
      & B+A     & RF    & .424 & [.279, .553] & .411 & 1.058 \\
      & B+S     & RF    & .403 & [.261, .537] & .418 & 1.055 \\
      & B+P+A   & Ridge & .455 & [.318, .580] & .469 & 1.026 \\
      & B+P+S   & Ridge & .480 & [.354, .593] & .500 & 1.006 \\
      & B+A+S   & Ridge & .479 & [.341, .597] & .414 & 1.056 \\
      & B+P+A+S & Ridge & \textbf{.494} & [.366, .607] & .492 & 1.011 \\
      & Claude Opus 5 (reference) & --- & .488 & --- & .399 & --- \\
\bottomrule
\end{tabular}
\vspace{2mm}
\footnotetext{
\footnotesize
\textit{Note.} $B$, $P$, $A$, and $S$ denote baseline features, LLM pairwise judgment, LLM absolute judgment, and LLM examinee readouts, respectively. Model = prediction model selected by training-set cross-validation within each configuration; RF = random forest. 95\% CI = percentile bootstrap confidence interval for the test-set Spearman correlation (2,000 resamples of test items). RMSE is not directly comparable across datasets.
}
\end{table}

Combining signal types produced small additional gains in rank-order performance. On USMLE, B+P+S achieved the highest Spearman correlation ($\rho = .447$), compared with .425 for B+P and .379 for B+S. Adding absolute judgment did not further improve rank-order performance, although the full B+P+A+S configuration achieved the highest Pearson correlation ($r = .443$) and the lowest RMSE (.278). On PANCE, B+P+S increased the Spearman correlation from .450 to .480 and B+P+A+S reached .494, whereas B+P+S achieved the highest Pearson correlation ($r = .500$) and the lowest RMSE (1.006). These gains were small relative to sampling uncertainty. In paired bootstrap comparisons, the gain of B+P+S over B+P was $\Delta\rho = .022$, 95\% CI [$-$.026, .071], on USMLE and $\Delta\rho = .030$, 95\% CI [$-$.015, .077], on PANCE, and the gain of B+P+A+S over B+P on PANCE was $\Delta\rho = .044$, 95\% CI [$-$.008, .099]. The gains were in the same direction in both datasets, but none of these intervals excluded zero. The best-performing configurations across both datasets used ridge regression.

On USMLE, the best results exceeded those reported for the winning system in the 2024 BEA NBME shared task ($\rho = .25$, $r = .27$, RMSE = .299). The highest Spearman correlations were also comparable to those of Claude Opus 5 on both USMLE (.447 vs. .443; $\Delta\rho = .004$, 95\% CI [$-$.102, .112]) and PANCE (.494 vs. .488; $\Delta\rho = .005$, 95\% CI [$-$.072, .087]). Overall, signals from open-weight LLMs, led by pairwise judgment, substantially improved difficulty prediction over baseline item features and achieved ranking performance comparable to direct judgments from a strong commercial model. The possibility of local deployment may make this approach particularly useful for unreleased or high-security test items.

\section{Discussion}
The main findings support a central idea of this study: signals that emerge from how LLMs interact with items can provide substantively useful information about item difficulty. Pairwise judgment was the most stable single LLM-derived difficulty signal across both datasets, while simulated-examinee signals also showed meaningful associations with empirical item difficulty. Adding simulated-examinee signals to pairwise judgments improved difficulty ranking on both USMLE and PANCE, but the gains were small and their confidence intervals included zero. The consistent direction across two independent datasets suggests that judgment-based and response-based signals may carry partly distinct information, but larger test sets will be needed to establish this. The results also substantially exceeded those of prior medical item difficulty prediction approaches that relied primarily on text representations. More broadly, when the target attribute cannot be read directly from the text itself, signals generated through model judgment and response behavior may offer a more promising source of predictive information than further improvements in text representation alone.

Pairwise comparisons generally produced stronger difficulty signals than absolute ratings, consistent with findings from the broader literature on LLM-based ranking. \citet{qin2024large} argued that the advantage of pairwise comparison comes not only from avoiding the need to calibrate absolute scores, but also from reducing a more complex global judgment task to a simpler local relative decision. This issue may be especially important in the present setting because absolute difficulty ratings must also be calibrated to a particular target examinee population. The prompts of the present study provided only minimal information about that population, such as ``USMLE examinees'' or ``PANCE examinees,'' without specifying ability levels, knowledge backgrounds, or anchor items with known empirical difficulty. Models may therefore lack the information needed to align their absolute judgments precisely with the difficulty scale of the actual examinee population. By contrast, pairwise comparison requires only a local relative judgment and is therefore less dependent on absolute scale calibration.

The comparison-density results further showed that pairwise judgment did not require a very large number of comparisons to produce a stable difficulty signal. For most models, approximately 10 to 15 comparisons per item were sufficient to reach about 95\% of the external predictive performance observed at $k=40$. This range is similar to previous studies. \citet{verhavert2019meta}, in a meta-analysis of 49 comparative judgment studies, found that approximately 10 to 14 comparisons per object were sufficient to achieve a scale separation reliability of about .70. \citet{kinnear2025comparative} reanalyzed 101 comparative judgment datasets from multiple disciplines and concluded that about 10 comparisons per object are generally sufficient. These criteria are not directly equivalent to the one used in the present study. Prior comparative judgment research has primarily focused on the reliability and stability of the resulting scale, whereas the present study evaluates the correlation between Bradley-Terry difficulty scores and empirical difficulty from real examinees. Still, both lines of evidence suggest rapidly diminishing returns once a moderate comparison density has been reached. At the same time, the present results show that increasing the number of comparisons cannot compensate for weak underlying judgment quality. In practice, it may therefore be more important to first select a model with adequate judgment quality and then determine the minimum comparison density needed for stable performance.

Results across judge models also showed that nominal parameter count was not a simple predictor of performance on the difficulty judgment task. Gemma 4 outperformed models with larger nominal parameter counts on both datasets, but this result should not be interpreted simply as evidence that smaller models outperform larger ones. Gemma 4 is a dense model, meaning that most of its approximately 31B parameters participate in each forward pass. In contrast, Nemotron 3 Super and GPT-OSS 120B use mixture-of-experts architectures, in which each token is routed to only a subset of experts. As a result, the number of parameters activated during a forward pass is much smaller than the total parameter count, at approximately 12.7B for Nemotron 3 Super and 5.1B for GPT-OSS 120B. At the same time, active parameter counts in mixture-of-experts models should not be directly compared with the total parameter counts of dense models, because the full set of experts still contributes to the overall capacity of the model. The marked difference between Gemma 4 and Gemma 3, together with the advantage of MedGemma over Gemma 3, suggests that model generation, training procedures, architecture, and domain adaptation may all contribute to performance on this task.

The simulated-examinee results showed a different pattern. The most informative response signals did not always come from the most capable models or those with the largest parameter counts. This finding is consistent with \citet{acquaye2026take}, who found in a study of simulated students for mathematics difficulty prediction that models with relatively weaker mathematical ability reproduced real student item performance more successfully than models with higher baseline accuracy. They suggested that highly capable models may be too proficient at solving the questions to reproduce realistic student error patterns. In the present study, a model that answers nearly all items correctly may similarly provide too little response variation to distinguish among items of different difficulty. At the other extreme, a model with insufficient task ability may produce more errors, but those errors may not correspond systematically to the difficulty ordering observed among real examinees. The most informative simulated examinee may therefore need both sufficient task competence and enough item-level response variation for its behavior to reflect differences in difficulty.

Although the observed correlations are insufficient to justify replacing traditional pretesting, LLM-derived difficulty estimates may serve as collateral information for item calibration when examinee response data are limited \citep{mislevy1988exploiting}. They may also support item bank management by flagging items predicted to be unusually easy or difficult for further review before administration.

\section{Limitations}

The study used the same relatively simple prompt across models within each task to reduce variation introduced by researcher-driven prompt optimization. However, identical prompts do not imply identical processing within different model systems. Commercial models may include system-level instructions or other forms of input processing that are not observable to researchers, while open-weight models typically provide more direct control over the inputs they receive. Future research could manipulate prompt detail directly to test whether richer descriptions of the target examinee population, examples, or model-specific prompt optimization alter the relative ordering of models.

The test sets were relatively small (the USMLE split was fixed by the shared task to allow direct comparison with prior results), so differences of a few hundredths between configurations should be interpreted cautiously.

The USMLE items were publicly released as part of the 2024 BEA shared task, and some of the LLMs evaluated in this study may have been exposed to these items during model development. The PANCE dataset provided an independent evaluation using empirical difficulty parameters that were not publicly available. Similar patterns across the two datasets provide additional support for the findings.

\backmatter

\bmhead{Acknowledgements}

The PANCE data were provided by the National Commission on Certification of Physician Assistants (NCCPA), which also approved the use of PANCE items with the commercial model reported in this study.

\bibliography{bibliography}

\clearpage
\begin{appendices}
\section{Supplementary Tables}
\setcounter{table}{0}
\renewcommand{\thetable}{A\arabic{table}}

\begin{table}[htbp]
\centering
\caption{Spearman Correlations Between LLM Examinee Readouts and Empirical Item Difficulty}
\label{tab:examinee_readouts}
\small
\begin{tabular}{lcccc}
\toprule
\textbf{Model} &
\textbf{$S_1$} &
\textbf{$S_2$} &
\textbf{$S_3$} &
\textbf{$S_4$} \\
\midrule
\multicolumn{5}{l}{\textit{USMLE} ($n=466$)} \\
Nemotron 3 Super 120B   & .276 & .251 & .205 & .274 \\
GPT-OSS 120B            & .021 & .124 & .001 & .020 \\
Llama 3.3 70B           & .278 & .093 & .265 & .277 \\
Llama 3.1 70B           & .291 & .132 & .277 & .289 \\
Gemma 4 31B             & .256 & .078 & .246 & .252 \\
Nemotron 3 Nano 30B     & .237 & \textbf{.270} & .037 & .234 \\
MedGemma 27B            & .283 & .147 & .283 & .275 \\
Gemma 3 27B             & .272 & .201 & .241 & .271 \\
Mistral Small 3.1 24B   & \textbf{.327} & .226 & \textbf{.287} & \textbf{.327} \\
GPT-OSS 20B             & .180 & .173 & .093 & .173 \\
Llama 3.1 8B            & .266 & .137 & .265 & .264 \\
Granite 3.3 8B          & .224 & .179 & .143 & .228 \\
Mistral 7B              & .129 & .140 & .002 & .124 \\
\addlinespace
\multicolumn{5}{l}{\textit{PANCE} ($n=370$)} \\
Nemotron 3 Super 120B   & .058 & .082 & .005 & .060 \\
GPT-OSS 120B            & .061 & .131 & .110 & .063 \\
Llama 3.3 70B           & .344 & .128 & .347 & .342 \\
Llama 3.1 70B           & \textbf{.386} & .135 & \textbf{.384} & \textbf{.384} \\
Gemma 4 31B             & .307 & .152 & .308 & .307 \\
Nemotron 3 Nano 30B     & .150 & .117 & .080 & .151 \\
MedGemma 27B            & .302 & .102 & .307 & .304 \\
Gemma 3 27B             & .275 & .144 & .291 & .279 \\
Mistral Small 3.1 24B   & .326 & .169 & .320 & .324 \\
GPT-OSS 20B             & .165 & .159 & .121 & .165 \\
Llama 3.1 8B            & .290 & .222 & .284 & .290 \\
Granite 3.3 8B          & .273 & \textbf{.269} & .156 & .272 \\
Mistral 7B              & .157 & .131 & .023 & .139 \\
\bottomrule
\end{tabular}
\vspace{2mm}
\footnotetext{
\footnotesize
\textit{Note.} Values are absolute training-set Spearman correlations.
$S_1$ = keyed-option probability; $S_2$ = binary correctness;
$S_3$ = response entropy; $S_4$ = top-two margin. Boldface indicates
the highest correlation within each readout and exam.
}
\end{table}

\begin{table}[htbp]
\centering
\caption{Selected LLM-Derived Signals and Training-Set Spearman Correlations}
\label{tab:signals}
\small
\begin{tabular}{llll}
\toprule
Dataset & Signal / Readout & Selected Model & $\rho$ \\
\midrule
USMLE & LLM pairwise judgment ($P$)        & Gemma 4 31B          & .424 \\
      & LLM absolute judgment ($A$)        & Gemma 4 31B          & .349 \\
      & Keyed-option probability ($S_1$)   & Mistral Small 3.1 24B & .327 \\
      & Binary correctness ($S_2$)         & Nemotron 3 Nano 30B  & .270 \\
      & Response entropy ($S_3$)           & Mistral Small 3.1 24B & .287 \\
      & Top-two margin ($S_4$)             & Mistral Small 3.1 24B & .327 \\
\midrule
PANCE & LLM pairwise judgment ($P$)        & Gemma 4 31B          & .439 \\
      & LLM absolute judgment ($A$)        & Gemma 4 31B          & .411 \\
      & Keyed-option probability ($S_1$)   & Llama 3.1 70B        & .386 \\
      & Binary correctness ($S_2$)         & Granite 3.3 8B       & .269 \\
      & Response entropy ($S_3$)           & Llama 3.1 70B        & .384 \\
      & Top-two margin ($S_4$)             & Llama 3.1 70B        & .384 \\
\bottomrule
\end{tabular}
\vspace{2mm}
\footnotetext{
\footnotesize
\textit{Note.} $\rho$ denotes training-set Spearman correlation. Within each
signal type or response readout, the selected model was the one with the largest
absolute correlation with empirical item difficulty. Signs are reported as absolute
values: keyed-option probability, binary correctness and top-two margin correlate
negatively with difficulty, response entropy positively.
}
\end{table}

\begin{table}[htbp]
\centering
\caption{Hyperparameter Grids for the Downstream Prediction Models}
\label{tab:learners}
\small
\begin{tabular}{lll}
\toprule
Model & Tuned & Fixed \\
\midrule
Ridge regression & $\alpha \in \{0.1, 1, 10, 100, 1000\}$ & --- \\
Random forest    & Features per split $\in \{.3, .7\}$ & 300 trees \\
                 & Min.\ leaf size $\in \{3, 10\}$ & \\
XGBoost          & Learning rate $\in \{0.03, 0.08\}$ & 300 trees; subsample = .8 \\
                 & Max.\ depth $\in \{2, 4\}$ & Column subsample = .8 \\
                 & & $\lambda = 1$; min.\ child weight = 5 \\
\bottomrule
\end{tabular}
\vspace{2mm}
\footnotetext{
\footnotesize
\textit{Note.} The grids yield 13 candidate settings (5 ridge, 4 random forest, 4 XGBoost). For each of the eight feature configurations, all 13 settings were evaluated by repeated five-fold cross-validation on the training set, and the setting with the highest mean cross-validated Spearman correlation was selected.
}
\end{table}

\clearpage

\section{Prompts}
\label{app:prompts}

The prompts used for each task are listed below. For the judgment tasks, \texttt{\{item\_text\}} denotes the item stem, response options, and keyed answer; in pairwise judgments, \texttt{\{first\_item\_text\}} and \texttt{\{second\_item\_text\}} denote the two items in their order of presentation. For the examinee task, the keyed answer was not shown.

\subsection*{Pairwise Judgment}

\begin{prompt}{System prompt (USMLE).}
You are a medical examinee taking the USMLE (United States Medical Licensing Examination).
You will see two items, A and B, each with its stem, options, and correct answer.
Some items refer to figures, photographs, graphs, or table column headers that examinees can see but that are not reproduced in the text you receive. Examinees have access to that material; its absence from your view is not itself a source of difficulty for them.
Judge which item is HARDER for examinees -- meaning fewer examinees answer it correctly.

Compare the two items directly: explain what specifically makes the harder one harder THAN THE OTHER, and where examinees are most likely to fail.

Output EXACTLY in this format:
<reasoning> your direct comparison of the two items </reasoning>
<harder> A or B </harder>
\end{prompt}

\begin{prompt}{System prompt (PANCE).}
You are a physician assistant examinee taking the PANCE (Physician Assistant National Certifying Examination).
You will see two items, A and B, each with its stem, options, and correct answer.
Some items refer to figures, photographs, graphs, or table column headers that examinees can see but that are not reproduced in the text you receive. Examinees have access to that material; its absence from your view is not itself a source of difficulty for them.
Judge which item is HARDER for examinees -- meaning fewer examinees answer it correctly.

Compare the two items directly: explain what specifically makes the harder one harder THAN THE OTHER, and where examinees are most likely to fail.

Output EXACTLY in this format:
<reasoning> your direct comparison of the two items </reasoning>
<harder> A or B </harder>
\end{prompt}

\begin{prompt}{User prompt.}
Question A
{first_item_text}

Question B
{second_item_text}

Which is harder? Follow the required output format.
\end{prompt}

\subsection*{Absolute Judgment}

\begin{prompt}{System prompt (USMLE).}
You are an experienced USMLE item writer and psychometrician. Examinees are students at US medical schools.
Difficulty is expressed as a percentile within a reference set of items from this examination. Percentile 0 marks the item that the largest share of examinees answered correctly; percentile 100 marks the item that the smallest share answered correctly. Items are spread evenly across the scale.
Some items refer to figures, photographs, graphs, or table column headers that examinees can see but that are not reproduced in the text you receive. Examinees have access to that material; its absence from your view is not itself a source of difficulty for them.

Estimate the difficulty percentile of the item shown.

Output EXACTLY in this format:
<reasoning> your analysis </reasoning>
<percentile> a whole number from 0 to 100 </percentile>
\end{prompt}

\begin{prompt}{System prompt (PANCE).}
You are an experienced PANCE item writer and psychometrician. Examinees are graduating students of US physician assistant programs.
Difficulty is expressed as a percentile within a reference set of items from this examination. Percentile 0 marks the item that the largest share of examinees answered correctly; percentile 100 marks the item that the smallest share answered correctly. Items are spread evenly across the scale.
Some items refer to figures, photographs, graphs, or table column headers that examinees can see but that are not reproduced in the text you receive. Examinees have access to that material; its absence from your view is not itself a source of difficulty for them.

Estimate the difficulty percentile of the item shown.

Output EXACTLY in this format:
<reasoning> your analysis </reasoning>
<percentile> a whole number from 0 to 100 </percentile>
\end{prompt}

\begin{prompt}{User prompt.}
{item_text}

Estimate the difficulty percentile. Follow the required output format.
\end{prompt}

\subsection*{Simulated Examinee}

\begin{prompt}{System prompt, non-reasoning models (USMLE).}
You are a medical examinee taking the USMLE (United States Medical Licensing Examination).
Answer the multiple-choice item below.
Reply with the single letter of the correct option and nothing else.
\end{prompt}

\begin{prompt}{System prompt, non-reasoning models (PANCE).}
You are a physician assistant examinee taking the PANCE (Physician Assistant National Certifying Examination).
Answer the multiple-choice item below.
Reply with the single letter of the correct option and nothing else.
\end{prompt}

\begin{prompt}{System prompt, reasoning models (USMLE).}
You are a medical examinee taking the USMLE (United States Medical Licensing Examination).
Answer the multiple-choice item below.
Think briefly, then end your reply with a line reading exactly:
Answer: X
where X is the letter of the correct option.
\end{prompt}

\begin{prompt}{System prompt, reasoning models (PANCE).}
You are a physician assistant examinee taking the PANCE (Physician Assistant National Certifying Examination).
Answer the multiple-choice item below.
Think briefly, then end your reply with a line reading exactly:
Answer: X
where X is the letter of the correct option.
\end{prompt}

\begin{prompt}{User prompt.}
{item_text}
\end{prompt}
\end{appendices}

\end{document}